\documentclass[preprint,12pt]{elsarticle}
\usepackage{hyperref,color,amssymb,amsmath}
\usepackage{graphicx}
\usepackage{subfigure}
\usepackage{float}
\usepackage{bbding}

\journal{Journal of *}

\begin{document}

\begin{frontmatter}

\title{End-to-end feature fusion siamese network for adaptive visual tracking}

\author[1]{Dongyan Guo}
\author[1]{Jun Wang}
\author[1]{Weixuan Zhao}
\author[1,2]{Ying Cui\corref{cor1}}
\cortext[cor1]{Corresponding author:
  Tel.: +86-13588864298;
}
\ead{cuiying@zjut.edu.cn}
\author[1]{Zhenhua Wang}
\author[1]{Shengyong Chen}

\address[1]{College of Computer Science and Technology, Zhejiang University of Technology, No. 288, Liuhe Road, Hangzhou, Zhejiang 310023, China}

\address[2]{Key Laboratory of Intelligent Perception and Systems for High-Dimensional Information of Ministry of Education, Nanjing University of Science and Technology, Nanjing, China}

\begin{abstract}
According to observations, different visual objects have different salient features in different scenarios. Even for the same object, its salient shape and appearance features may change greatly from time to time in a long-term tracking task. Motivated by them, we proposed an end-to-end feature fusion framework based on Siamese network, named FF-Siam, which can effectively fuse different features for adaptive visual tracking. The framework consists of four layers. A feature extraction layer is designed to extract the different features of the target region and search region. The extracted features are then put into a weight generation layer to obtain the channel weights, which indicate the importance of different feature channels. Both features and the channel weights are utilized in a template generation layer to generate a discriminative template. Finally, the corresponding response maps created by the convolution of the search region features and the template are applied with a fusion layer to obtain the final response map for locating the target. Experimental results demonstrate that the proposed framework achieves state-of-the-art performance on the popular Temple-Color, OTB50 and UAV123 benchmarks.
\end{abstract}


\end{frontmatter}

\section{Introduction}
Visual object tracking is one of the hotspots in computer vision. Object tracking is widely employed in many real-world visual applications, such as autonomous driving, video surveillance, human-computer interaction, \emph{etc..} The task of object tracking is estimating the trajectory of an object in an image sequence. However, the only knowledge about the object is the target location in the first frame. The lack of priori knowledge renders the task challenging. Besides, the problem is also challenged because of many influences such as illumination variations, scale variations, non-rigid deformations, fast motion, background clutters, motion blur and partial occlusions. 

In recent years, correlation filter based methods have shown excellent performance on object tracking benchmarks \cite{wu2015object}. However, most of these approaches only use hand-crafted appearance features to present the tracking target, which cannot get satisfactory performance in many applications including occlusions \cite{danelljan2014accurate,danelljan2017discriminative,zhang2014meem} and background clutters \cite{bertinetto2016staple,li2014scale,zhang2014meem}. On the other hand, deep neural networks can achieve excellent performance in many applications where enough priori knowledge of the target can be obtained for training powerful models. However, the lack of priori knowledge is the chief challenge for applying deep neural networks to object tracking task. Moreover, on-line updating is very time consuming, especially when a large number of parameters are involved. It is therefore crucial to balance the tracking accuracy and speed.

One possible way to solve the aforementioned problems is to train deep neural networks model off-line. Some existing works adapt a pre-trained model for the target to get CNN (Convolutional Neural Network) features \cite{danelljan2015convolutional,danelljan2016beyond,ma2015hierarchical}. Though the pre-trained model saves the time of online updating, its fixed metric prevents the learning strategy from exploiting the sense-specific cues which are significant for discrimination. Some approaches use Siamese CNN architecture, an offline adaptation network \cite{bertinetto2016fully,chen2017once,held2016learning,leal2016learning,tao2016siamese}, which achieved the state-of-the-art performance. Besides, some research works combining on-line learning method with pre-trained CNN features has obtained successful improvement.

\begin{figure}
	\centering
	\includegraphics[width = \linewidth]{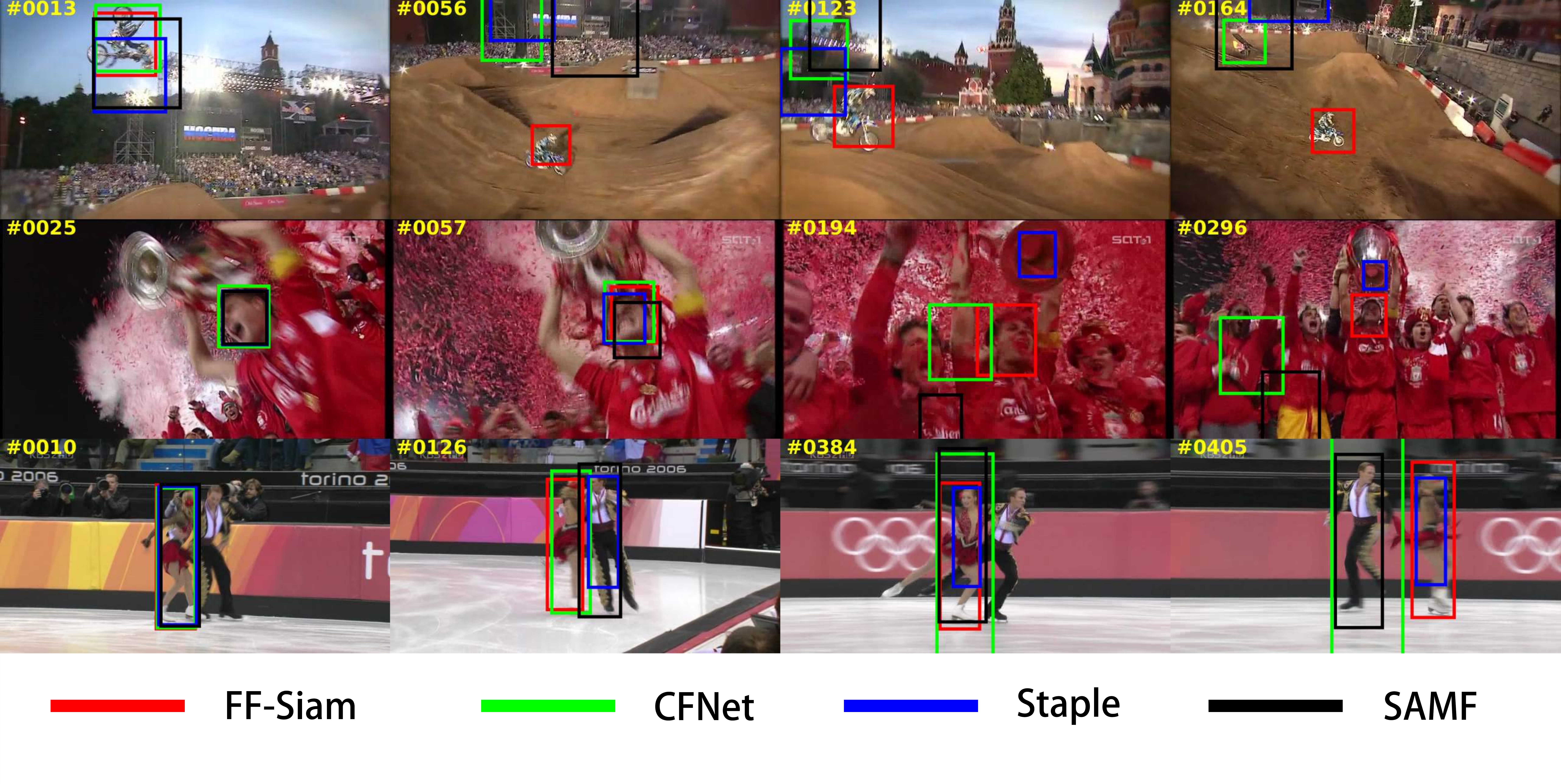}
	\caption{A qualitative comparison of the proposed approach(FF-Siam) with other three state-of-the-art approaches on three example sequences: Motor Rolling (as shown in the top row), Soccer (as shown in the middle row) and skating2 (as shown in the bottom row). These examples include the following cases: scale variation, occlusion, deformation, fast motion, out-of-plane rotation, in-plane rotation and background clutters. Our approach achieves superior results in these scenarios.}
	\label{fig:excellentperformance}
\end{figure}
When using a single features, both online and offline methods can give poor tracking results.
For example, HOG (Histogram of Orientated Gradient) is a general feature which has been employed in many state-of-the-art methods \cite{bertinetto2016staple,danelljan2014accurate,danelljan2017discriminative,henriques2015high}. But the HOG feature based methods have a common drawback: they are sensitive to large deformation. The trackers perform poorly when the object appearance change rapidly. The CNN features are powerful in image representations and stable for deformation. It turns out that as long as there are enough diverse training samples, the CNN features can achieve excellent performance even in scenarios with large object variations and background clutters. The relatively shortcoming is that if the training samples are not enough and lacking some kinds of scenes, the performance of CNN will drop very fast. However, fusing different features which are complementary to each other in the object tracking task is a proper way to solve this problem.

In this paper, we find that combining the CNN with hand-crafted appearance features like HOG can improve the universality of the tracker (see Figure 1) and propose a feature fusion siamese network called FF-Siam for adaptive visual tracking. We find that different feature channels have their own advantages for different scenarios, then a channel attention mechanism is introduced to combine the different channels of the same feature differently according to scenarios, which makes the tracker more discriminative. The features fused in this paper are CNNs and HOGs. However, it must be noted that the proposed approach is actually a general feature fusion network framework which can be easily extended to fuse other features. In the proposed network, we combine the CNN and Correlation Filter to generate a discriminative template for CNN and HOG features respectively, which can be used to compute CNN and HOG response maps. Thousands of parameters have been trained through the Siamese Network framework to improve the CNN and HOG features fusion results, which in turn benefit the tracking performance. We  show the architecture of the proposed network in Figure 2. The network architecture can be divided into four parts. First, we design a feature extraction layer to extract different features of the target region and the search region. The target region is a target-centered image patch cropped from the previous frame and the search region is the search area cropped from the current frame according the bounding-box in the previous frame. Second, the extracted features are input into the weight generation layer to obtain the channel weights. Third, the template generation layer utilizes the features and the channel weights to generate a corresponding template. Finally, the corresponding response maps achieved by the convolution of the search region features and the discriminative templates are applied with a fusion layer to obtain the final response map, which is used to locate the target.

Most existing feature fusion methods specify the fusion parameters manually, while our fusion parameters are learned end-to-end. By reducing manual preprocessing and subsequent processing, the proposed network gives the tracking model more freedom to be adjusted adaptively based on the current tracking sequence. In order to validate the effectiveness and the efficiency of the proposed approach, comprehensive evaluations and comparisons with other state-of-the-art trackers are conducted on the popular Temple-Color, OTB and UAV123 benchmarks. Experimental results demonstrate that the proposed approach is robust against complex backgrounds, shape deformation, color variation, and achieves state-of-the-art performance on these benchmarks.
\begin{figure}
	\centering
	\includegraphics[width = \textwidth]{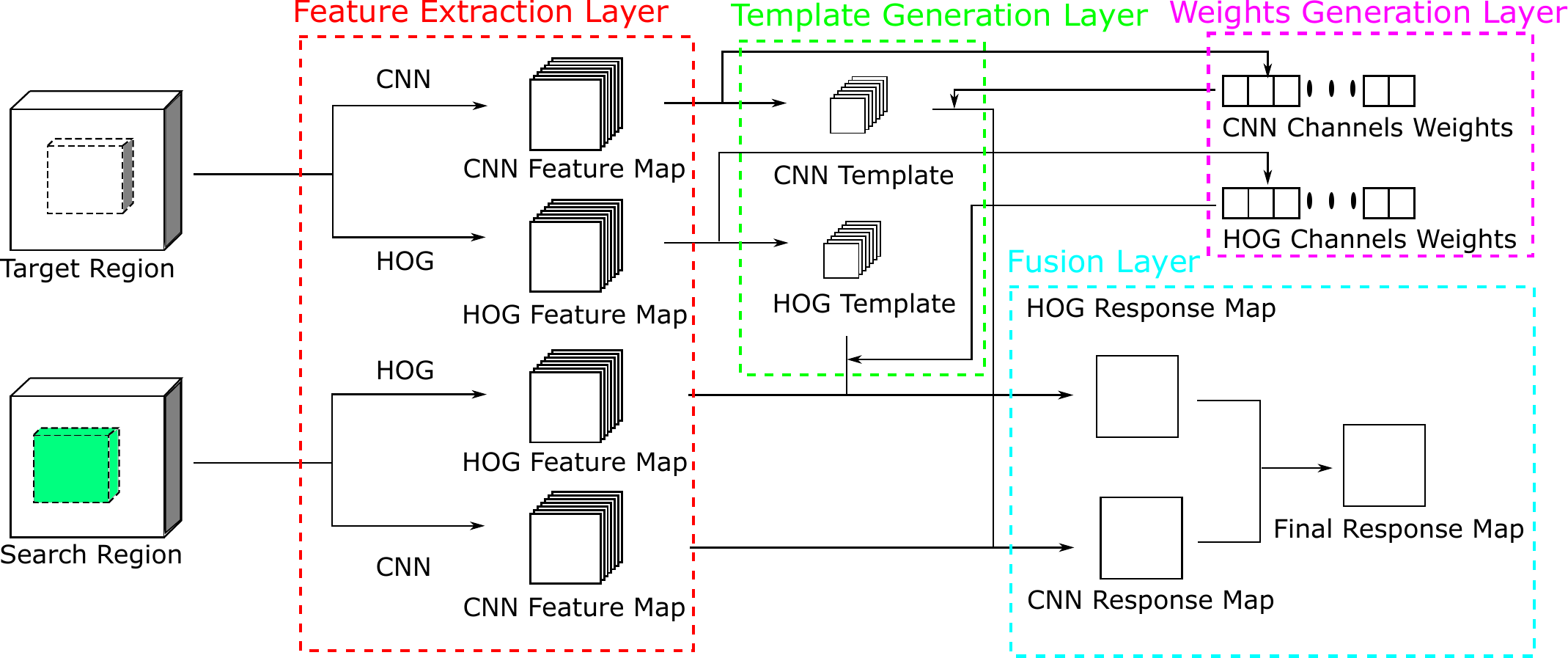}
	\caption{The architecture of our proposed network. }
	\label{fig:networkarchitecture}
\end{figure}
\section{Related Work}
\subsection{Siamese Network Based Trackers}
 With the development of deep learning, methods based on deep learning are developed rapidly in the visual object tracking. Recent works have focused on learning universal object descriptors to solve the tracking problems. The methods \cite{bertinetto2016fully,chen2017once,held2016learning,leal2016learning,tao2016siamese,valmadre2017end} based on the Siamese CNN architecture trained off-line can take advantage of information presented in numerous training images. The basic Siamese Network architecture \cite{bromley1993signature} was first proposed by Bromley et al., which is a simple but powerful network. The network has excellent performance in discriminating whether or not the same object in two different image patches. It has been applied in many visual task, such as face verification\cite{chopra2005learning,taigman2014deepface} and local image patch comparison \cite{han2015matchnet,zagoruyko2015learning}. In recent years, Siamese Network has also been widely used in the field of object tracking. The tracker SiamFC\cite{bertinetto2016fully} is an end-to-end fully-convolutional architecture. The network architecture of SiamFC \cite{bertinetto2016fully} is very simple as it only consists of several convolutional layers, but it performes well on multiple benchmarks. The tracker SINT \cite{tao2016siamese} applies the Siamese deep neural network to learn a matching function which is used to find the most similar patch between a new frame and the first frame. The design of the two branches of the network is inspired by AlexNet \cite{krizhevsky2012imagenet} and VGGNet. The tracker CFNet \cite{valmadre2017end} is based on SiamFC, which introduces the CF into the network. The CFNet uses only two convolutional layers and reach state-of-the-art performance.
\subsection{Feature Fusion}
It is difficult to achieve satisfactory tracking results with single feature descriptor. Many works prove that combining multiple estimates can improve tracking results. The tracker in \cite{wang2014ensemble} adopts an ensemble-based structure. It uses a factorial HMM (Hidden Markov Model) to combine the results of five independent trackers. Even for the same video sequence, the reliability of the results obtained by different tracking algorithms is different. Therefore the tracker in \cite{wang2014ensemble} uses a factorial HMM to combine those results and get a better final result. The tracker in \cite{kwon2010visual} is based on a visual tracking decomposition structure. It decomposes the tracking task into multiple different sub-tasks. The results of multiple different basic observation models and motion models are integrated to achieve the task. Finally, it adopts Markov Chain Monte Carlo (MCMC) framework to contact these basic models to reach the tracking task. The tracker MEEM \cite{zhang2014meem} adopts a multi-expert restoration scheme. Unlike those trackers mentioned above, it integrates different types of trackers, MEEM stores a collection of past models and adopts the results of the past models to achieve the tracking task. For each frame, the tracker can obtain an evaluation result equals to the number of storage models. By using the loss entropy function, the optimal one is selected from these results. Bertinetto et al. find that the performance of template models based trackers like DSST \cite{danelljan2014accurate} is unsatisfactory when the object changes rapidly. When the color of the target is very similar to the color of the background, the color-based models like DAT \cite{possegger2015defense} is useless to distinguish the object from the background. They propose the tracker Staple \cite{bertinetto2016staple} which combines HOG and color histogram together to make up for the defect of the two features, so it can make the tracker robust to target deformations and background clutters. The tracker in \cite{chen2017once} combines the shallow features and deep features. The deep features are useful to distinguish the object from the background and the shallow feature is useful to represent the appearance of an object. 

The above trackers which only fuse CNN features with CNN features or fuse hand-crafted features with hand-crafted features cannot describe target well. For example, the hand-crafted features lack semantic information and the CNN features are insensitive to deformation.

To design a universal feature descriptor, we design an end-to-end feature fusion network to adaptively combine CNN features and hand-crafted features. The proposed network effectively solves the parameter learning problem in feature fusion and improves the adaptiveness and universality of the object tracker.

\section{The Proposed Approach}
We briefly introduce our proposed network framework in section 3.1. Then, the usage of the Correlation Filter to generate CNN and HOG template is explained in section 3.2; The detailed architecture of the channel-weights generation layer is represented in section 3.3. At last, we illustrate the use of the fusion model for object tracking in section 3.4.

\subsection{Feature Fusion framework}
Most online CNN-based trackers cannot work in real-time because of the expensive computation. In order to get a good balance of efficiency and performance, it is a good choice to train the model off-line. SiamFC treats object tracking as a similarity learning problem and trains a model off-line to achieve an excellent performance in real-time.

The end-to-end network we proposed to fuse CNN features and hand-crafted features is based on SiamFC. For training, the inputs of the network are pairs of image patches ($x'$, $y'$). The image $x'$ represents the object of interest in one frame of an image sequence. The image $y'$ represents the object search area in another frame which is randomly chosen in the same image sequence and the size of $y'$ is the same as $x'$. By training these pairs of images, the generalization ability of the models can be improved significantly.

HOG and CNN features are extracted from the two inputs respectively and each feature can obtain a response map by correlation filter operation. In order to make better use of these response maps, we set different weights, and fuse them together to obtain a new response map. The fusion approach can make full use of the advantages of the two features and make up for the deficiency between the two features. Therefore the tracking performance can be improved. Here we utilize the function $f_c$ and $f_h$ to extract CNN and HOG features from an image respectively. We apply the function $\phi_c$ and $\phi_h$ to generate the channel weights. A pair of image patches yield four feature maps (two CNN feature maps $f_c(x')$, $f_c(y')$ and two HOG feature maps $f_h(x')$, $f_h(y')$) which create two response maps $g_c(x',y')$, $g_h(x',y')$ after cross-correlated operation:

\begin{equation}
g_c(x',y') = (\phi_c(f_c(x')) \otimes W(f_c(x'))) \star f_c(y'), \label{eq1}
\end{equation}
\begin{equation}
g_h(x',y') = (\phi_h(f_h(x')) \otimes W(f_h(x'))) \star f_h(y'), \label{eq2}
\end{equation}
where the function $W$ is represented to get the optimal template w, and Eq. \ref{eq1} gets the CNN feature response map while eq. \ref{eq2} gets the HOG feature response map, then the two response maps fused as below:
\begin{equation}
g(x',y') = M_\rho(g_c, g_h), \label{eq3}
\end{equation}
in order to make the response map more suitable for logistic regression, the scale $s$ and bias $b$ are added into $m(x',y')$ to get the function $M(x',y')$,
\begin{equation}
M(x',y') = sm(x',y') + b, \label{eq4}
\end{equation}
here, the distance between the maximum location and the center of the response map is related to the offset of the target and the image center.

In order to ensure real-time tracking, the network is trained offline. The training images sampled of the network are millions of random pair (${x'}_i$ , ${y'}_i$). Each image pair has a spatial map of label information which is composed of ${x'}_i$ . The label is represents whether the pixel point is belongs to the ground truth or not.
\begin{equation}
L_i(r,c) = 
\begin{cases}
-1, &$not belong to ground truth; $\\
1,  &$belong to ground truth,$
\end{cases} \label{eq5}  
\end{equation}
in Eq. \ref{eq5}, $r$ and $c$ represent the row number and column number of the spatial map. The purpose of the network training is to minimize the element-wise logistic loss function:
\begin{equation}
\arg\min\sum_{i}\ell(g(x',y'), L_i). \label{eq6}
\end{equation}

Feature fusion is achieved by fusing the response maps which obtained by different features respectively. In this way, we can select better response map which is equivalent to select a better feature for tracking. The fusion approach can be represented as :
\begin{equation}
m(x',y') = \sum_{d=1}^{D}g_d(x',y') * k_d, \label{eq7}
\end{equation}
where D is the amount of the feature maps, $k_d$ is the fusion kernel trained by our
network.

\subsection{Correlation Filter}
The structure of full convolution network in SiamFC is effective for tracking. However, the CNN feature, as a semantic feature proposed for image classification, is insensitive to the apparent changes of the object. In other words, it lacks the specific target discriminant information. The Correlation Filter (CF) is a method to describe the relationship between two images, which can discriminate the relationship of image and image transformation well. Therefore, the discrimination of tracker can be improved by combining CNN and CF. The problem of solving the correlation filter template is equivalent to solving the ridge regression problem. In the following, the correlation filter template is denoted as $w$, and $x \in \mathbb{R}^{m\times m\times K}$ is a K-channel feature image, $y \in \mathbb{R}^{m\times m}$ is the desired response map. Under a least-squares Correlation Filter formulation, the problem can be represented as:
\begin{equation}
\arg\min_w\|w\star x - y {\|}^2 + \|w{\|}^2, \label{eq8}
\end{equation}
where symbol $\star$ denotes the circular cross-correlation, and the second term is the quadratic regularization which is added to avoid overfitting. By solving the partial derivative of the above formula to $w$ equals zero, the solution of the above formula is:
\begin{equation}
\hat{w} = \frac{\hat{y}^* \circ \hat{x}}{(\hat{x}^* \circ \hat{x}) + \lambda}, \label{eq9}
\end{equation}
In order to apply back-propagation to correlation filtering, the filter is gained like CFNet:
\begin{equation}
\hat{w} = \frac{\hat{y}^* \circ \hat{z}}{(\hat{z}^* \circ \hat{z}) + \lambda}, \label{eq10}
\end{equation}
where $z$ is the target region feature whose size is the same as the search patch. Then the filter is cropped to obtain the target areas, and the response map is obtained by convoluting the target area with the search area.


%

\subsection{Channels Weights Generation}
\begin{figure}
	\centering
	\includegraphics[width = \linewidth]{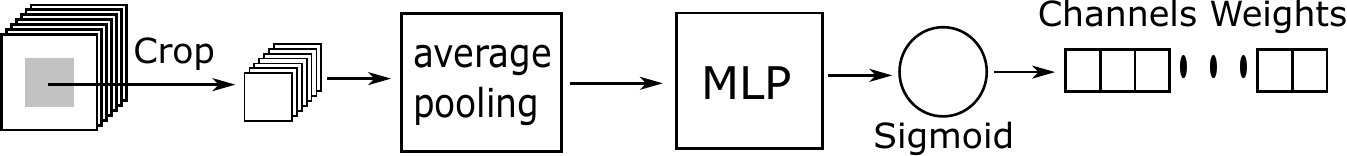}
	\caption{The detailed architecture of the channels weights generation layer}
	\label{fig:channels_weights_arc}
\end{figure}

Attention mechanism is first introduced in neuroscience by Olshausen et al.\cite{olshausen1994neurobiological}. In recent years, attention mechanism combined with deep learning has been popular in image classification \cite{hu2018squeeze,jaderberg2015spatial}, natural language processing \cite{bahdanau2015neural,luong2015effective}, speech recognition\cite{shan2017attention,chorowski2014end} and so on. In order to enhance the tracking robustness, here we introduce attention mechanism to the tracking framework for adaptive feature selection.  Different feature channels have different influences when track on different scenes, and the channel weights indicate different importance of different channels. For instance, when the object is moving fast, more attention should be paid on features which are insensitive to motion blur. Therefore, by increasing the weights of more stable feature channels, the adaptability of the algorithm can be improved. The channel attention architecture is simple to be implemented and needs only a few training parameters, but it improves the performance of the tracker significantly without consuming a lot of computation. The detailed architecture of the channel weight generation layer is shown in Fig. \ref{fig:channels_weights_arc}. 

The inputs of this part are features extracted from the image $x'$. The cropping operation takes out the middle area of the feature maps. Then the corresponding channel weights are obtained by passing features through a multilayer perceptron (MLP) which consists of two fully connected layers. The number of channel weights is equal to the number of feature map channels. The Channel weights make the combination of different channels with the same feature much closer, and different channels have different weight coefficients to distinguish different channel importances. To all kinds of the feature extraction, we apply the channel weight generation layer to improve the quality of features.

\subsection{Adaptive Visual Object Tracking Algorithm}
The input of the model is a pair of image patches consisting of the target region in the previous frame and the search region in the current frame. The search region is extracted as a sub-window centered on the previous estimated position and its size is four times of the object. The output of the model is the fusion response map. The location of maximum value of the response is corresponding to the center of the object. To solve the scale variations, we input the search area of three scales, and take the maximum of these response maps as the result.
Although the model is trained offline, we find that the updating strategies using online learning can improve the tracking performance. Taking a pair of image patches as the input, two new templates $W(f_c(x'))$ and $W(f_h(x'))$ are output by the model. The approach to update old feature template with new feature template is shown in Eq. \ref{eq11}.
\begin{equation}
\begin{split}
Temp_{c,new}&=(1 - \eta_c)Temp_{c,pre} + \eta_cW(f_c(x')),    \\
Temp_{h,new}&=(1 - \eta_h)Temp_{h,pre} + \eta_cW(f_h(x')),
\end{split} 
\label{eq11}
\end{equation}
where the parameter $\eta$ represents the learning rate of the template.
\section{Experiments}
We evaluate our proposed FF-Siam network by performing contrast experiments on three benchmarks: Temple-Color \cite{liang2015encoding}, UAV123 \cite{mueller2016benchmark}, and OTB50 \cite{wu2013online,wu2015object}. The purpose here is to evaluate the effect of using our network to train parameters for feature fusion. First, we compare the effects of different convolutional layer depths on the tracking performance. Then, we compare our approach with some state-of-the-art trackers on benchmarks. Bounding box overlap ratio and center location error are two metrics to evaluate the tracker performance. The bounding box overlap ratio is defined to measure the bounding boxes overlap of ground truth $R_{gt}$ and the tracker’s predict result $R_t$, which is
\begin{equation}
S(\sigma_{over}) = \frac{R_{gt}\cap R_t}{R_{gt} \cup R_t}\geq \sigma_{over}, \label{12}
\end{equation}
where $\sigma_{over}$ is the threshold of bounding box overlap ratio which range is [0,1]. the center location error is defined as the Euclidean distance of the bounding box centers between ground truth $P_{gt}$ and the trackers predict result $P_t$, which is
\begin{equation}
P(\sigma_{succ}) = \|P_{gt} - P_{t} \| \leq \sigma_{succ}, \label{13}
\end{equation}
where the $\sigma_{succ}$ is the threshold of center location error which unit is pixel.

\subsection{Implementation Details}
CNN feature: The network we use to extract CNN features is based on SiamFC. However, in order to combining the correlation filter method, the size of the input training samples is the same as that of the test samples, which is fixed to $255×255$. Meanwhile, the output of feature size depends on the depth of the network. For example, when two convolution layers are used, the output feature size is $57×57×32$, and for three-layer convolution, the output feature size is $53×53×32$. Then we crop the output target region features to ensure the dimension of the response map is $33×33$.

HOG feature: The HOG feature of target region branch and search region branch have the same size. The input size of images is $255×255$ and the output feature size is fixed to $62×62×31$. Then we crop a patch of $30×30$ in the center of target region features for each channel. Therefore, the dimension of the response map is $33×33$. 

Attention module: We apply attention network to all types of features. The features are cropped to ensure the input without too much background, and the following pooling layer is changed to average pooling. The number of output neurons of MLP is the same as the number of channels, and the activation function in MLP is ReLU. Finally, we use a Sigmoid function added with bias to obtain the weight of each channel.

Training: All the parameters in this network are trained in an end-to-end manner on the ILSVRC-2015 video dataset. The parameters in the search region branch are the same as the target region branch. Therefore, we only need to train the parameters in target region branch. We train the network for 100 epochs, the initial value of learning rate is 0.01, we decreases learning rate by 0.9 every one epoch in the top 50 epoch and remained unchanged in the latter epoch.

\subsection{ Evaluation Of Different Convolutional Layer Depths}

\begin{figure}
	\centering
	\subfigure[conv1]{
		\label{Figcompare1}
		\includegraphics[width=0.3\textwidth,height= 3.5cm]{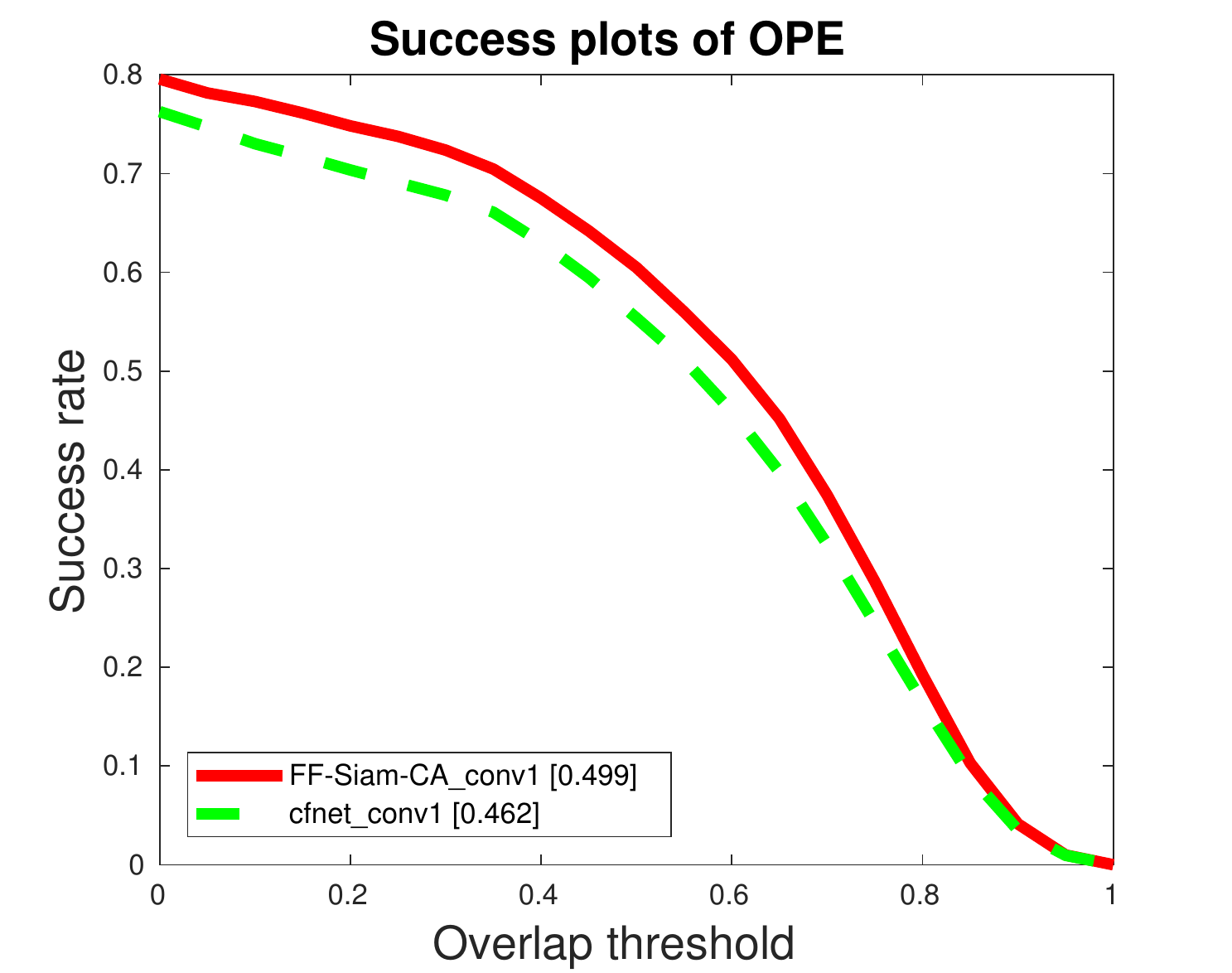}}
	\subfigure[conv2]{
		\label{Figcompare2}
		\includegraphics[width=0.3\textwidth,height=3.5cm]{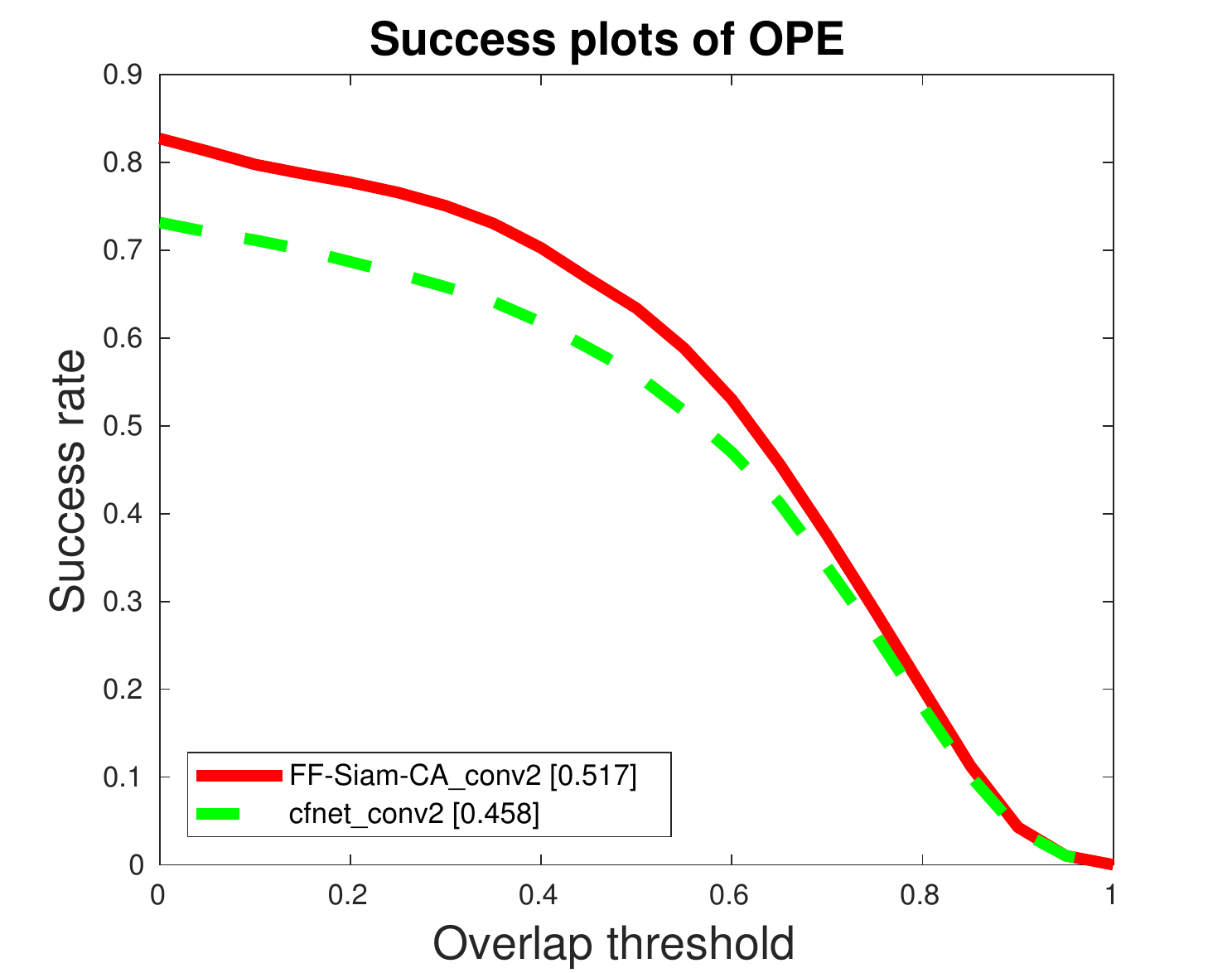}}
	\subfigure[conv5]{
		\label{Figcompare5}
		\includegraphics[width=0.3\textwidth,height=3.5cm]{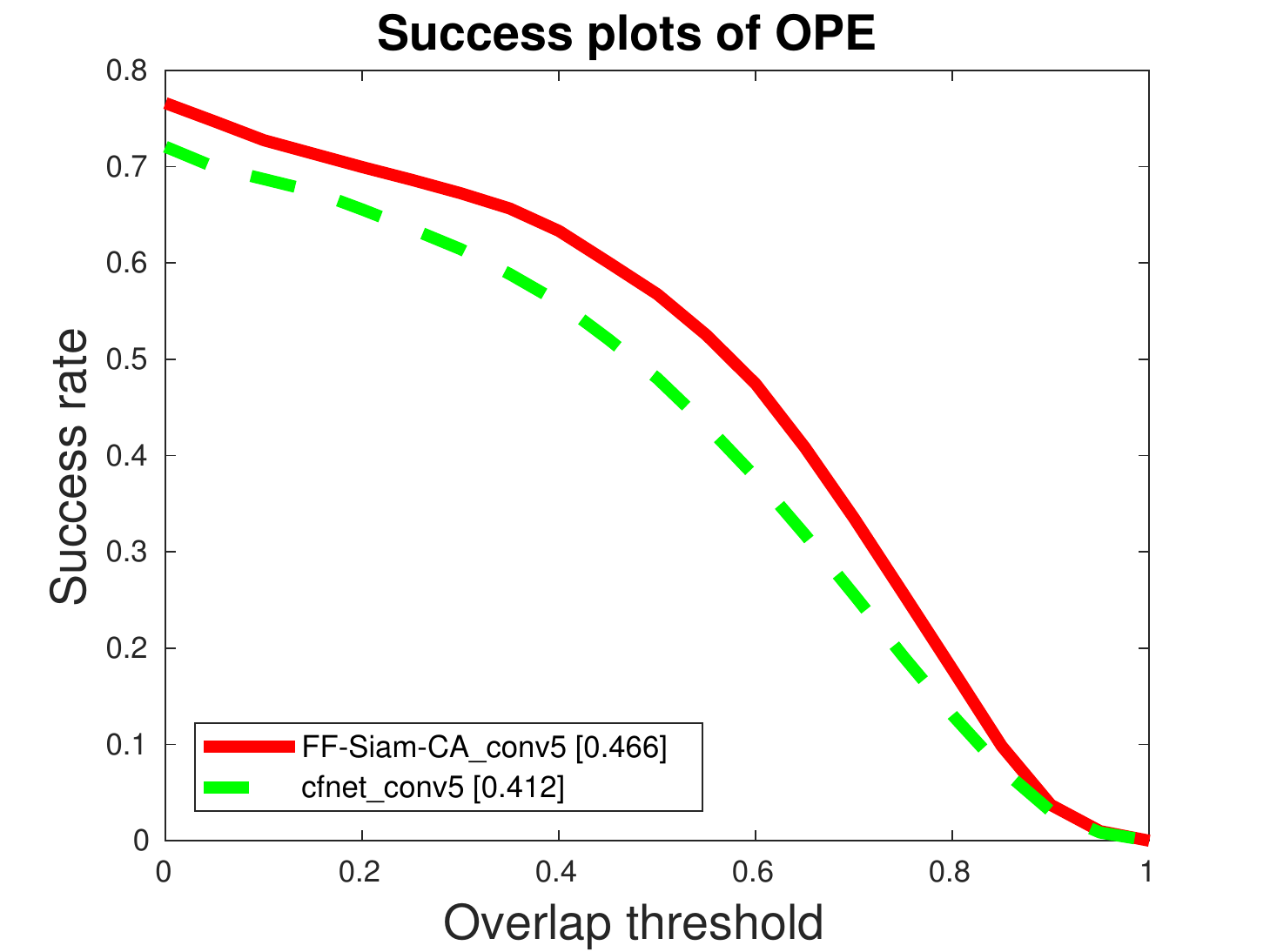}}
	\caption{Success rates of rectangle overlap for different convolutional layers on the validation dataset Temple-Color.}
	\label{Figcompare}
\end{figure}

\begin{figure}
	\centering
	\includegraphics[width = 0.85\linewidth,height=7.5cm]{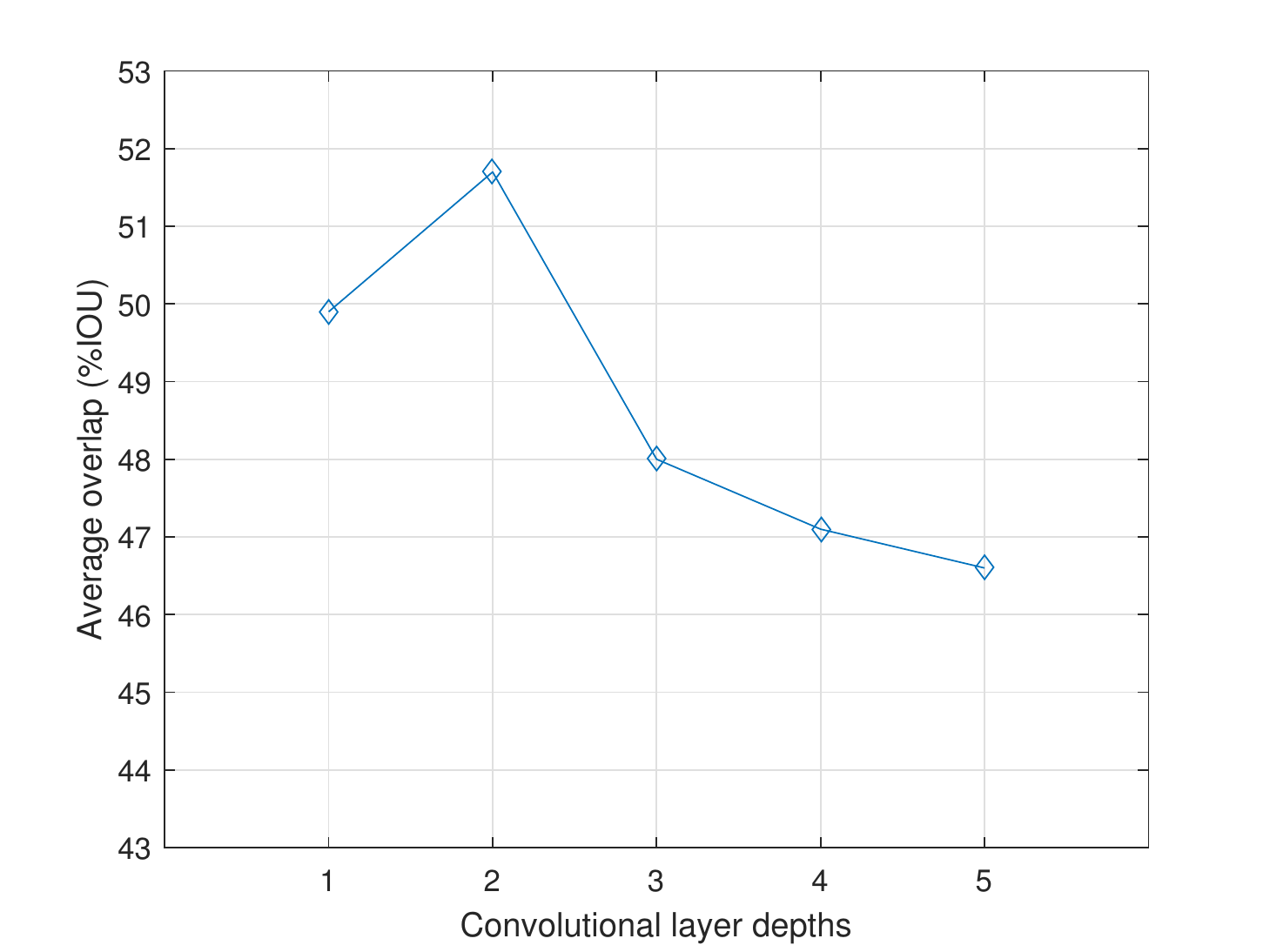}
	\caption{The accuracy with different convolutional layer depths of our approach.}
	\label{differelayer}
\end{figure}
In this part, we use the bounding box overlap ratio to evaluate the trackers. The success rate is calculated as the percentage of frames with intersection-over-union (IOU) exceeding a threshold. Temple-Color is the validation dataset in this part. Since we can only get the model of CFNet [46] using Conv-1, Conv-2 and Conv-5, the proposed approach uses the same convolutional layers for comparison. As shown in Figure 4, 4(a) represent the results of the first layer of CNN feature used in our method, while 4(b) and 4(c) correspond to the results of layer 2 and layer 5 respectively. In 4(a), the red curve describes the success rate corresponding to different thresholds in the proposed approach while the green curve describes that of CFNet. The legend represents the average success rate of all threshold. It can be seen that the result of our approach is significantly better than CFNet, which suggests that the combination of the CNN and HOG features improves the performance of the tracker. Besides, when the threshold is small, the improved performance of Conv-2 CNN features is better than that of Conv-5 CNN features, but as the threshold increases, the performance improvement of Conv-2 CNN features decreases faster, which indicates that the deeper the CNN features, the less apparent feature has, and the hand-crafted features can make up for this defect. To show the tracking results of HOG fused with different convolutional layers, we compare the fusion results of different CNN features and HOG features in Figure 5. Here the X-axis is the depth of the CNN features we used and the Y-axis is the average success rate. We find that Conv-2 achieves better results. When more convolutional layers are added, it seems to be redundant, which indicates that the Conv-2 is more suitable to be fused with hand-crafted features. It is because when we combine CNN with correlation filter method, the Conv-2 features performs better. Meanwhile, the Conv-2 features contain semantic information, it is suitable to fuse with HOG features.

\subsection{Comparisons With State-of-the-art Methods}
We compare the proposed approach with 13 state-of-the-art trackers: KCF \cite{henriques2015high}, Staple \cite{bertinetto2016staple}, SAMF \cite{li2014scale}, SiameFC \cite{bertinetto2016fully}, CFNet \cite{valmadre2017end}, MEEM \cite{zhang2014meem}, SRDCF \cite{danelljan2015learning}, DSST \cite{danelljan2014accurate}, DAT \cite{possegger2015defense}, ACT \cite{danelljan2014adaptive}, TGPR \cite{gao2014transfer}, KCFDP \cite{huang122015enable} and fDSST \cite{danelljan2017discriminative}. The success rate and precision rate are used to evaluate these trackers in the experiments. 

\textbf{Datasets:} UAV123 is a large dataset which is captured from low-altitude UAVs. The dataset consists of sequences from an aerial viewpoint, containing a total of 123 video sequences and more than 110K frames. OTB50 is the most challenging dataset of OTB datasets, therefore the experiments are only done on OTB50. The dataset contains 50 video sequences.

\begin{figure}
	\centering 
	\subfigure[UAV123]{\label{fig:OTB50AUC}
		\includegraphics[width=0.4\linewidth,height=4.2cm]{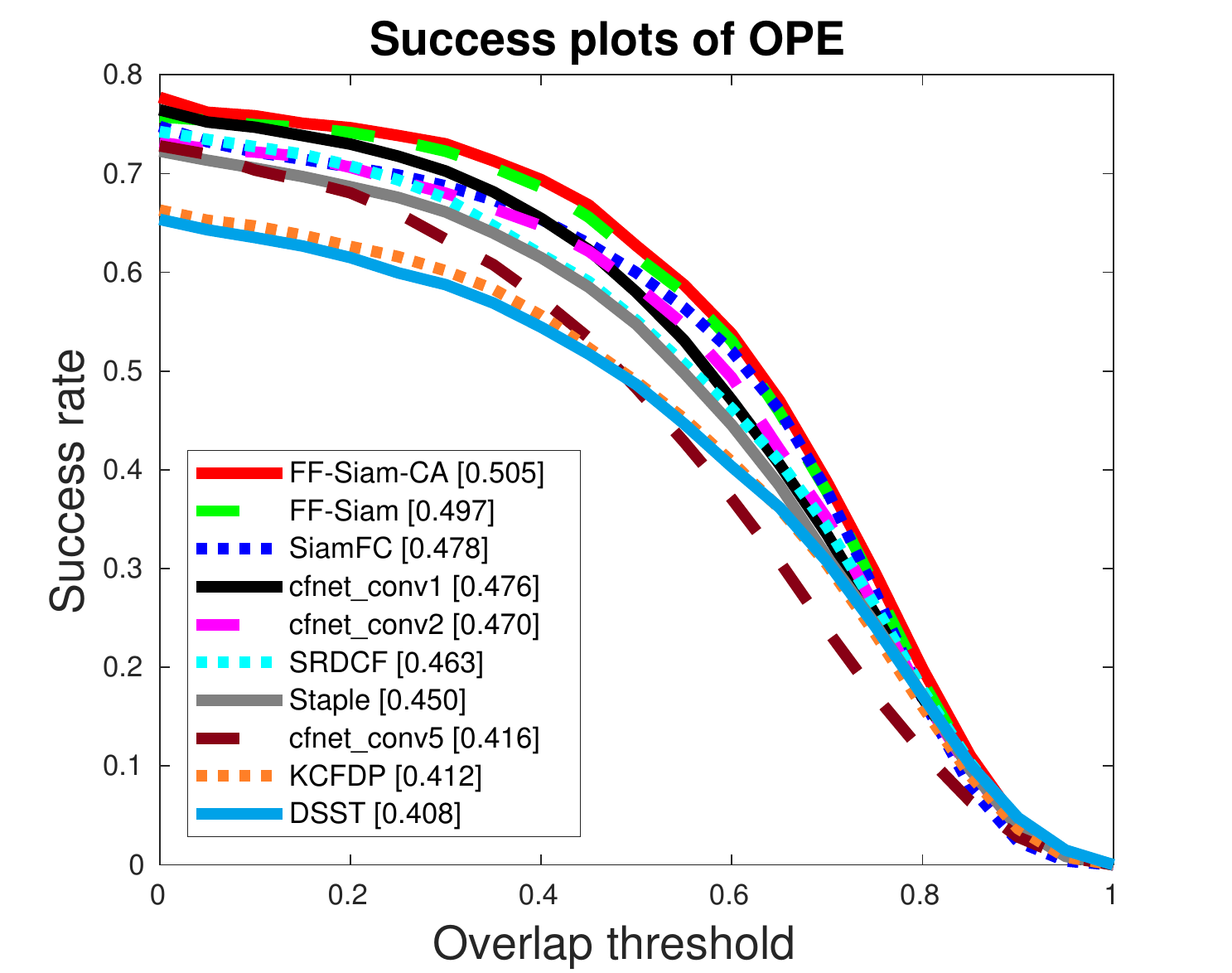}}
	\hspace{0.01\linewidth}
	\subfigure[UAV123]{\label{fig:OTB50pre}
		\includegraphics[width=0.4\linewidth,height=4.2cm]{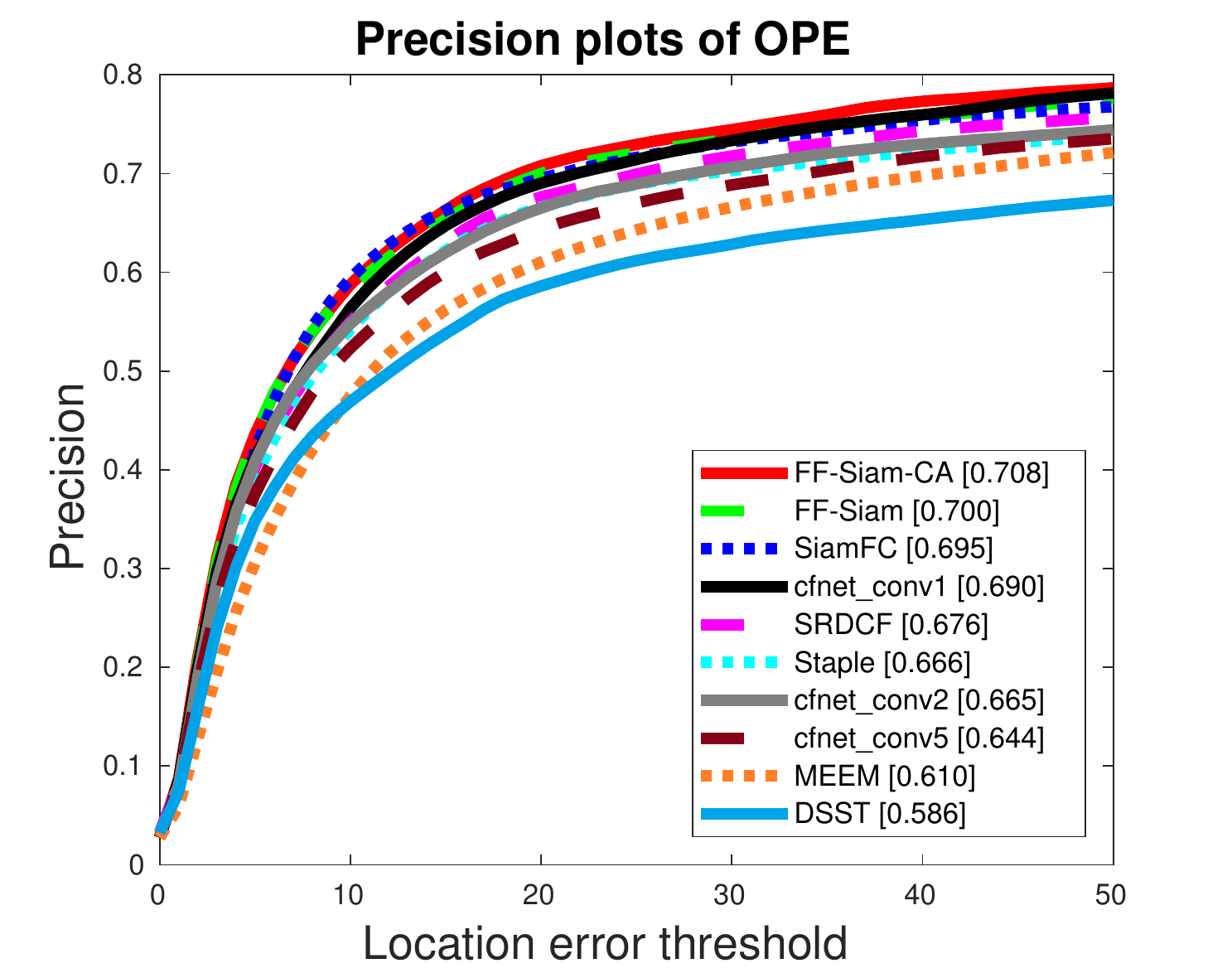}}
	\vfill
	\subfigure[OTB50]{\label{fig:UAV123AUC}
		\includegraphics[width=0.4\linewidth,height=4.2cm]{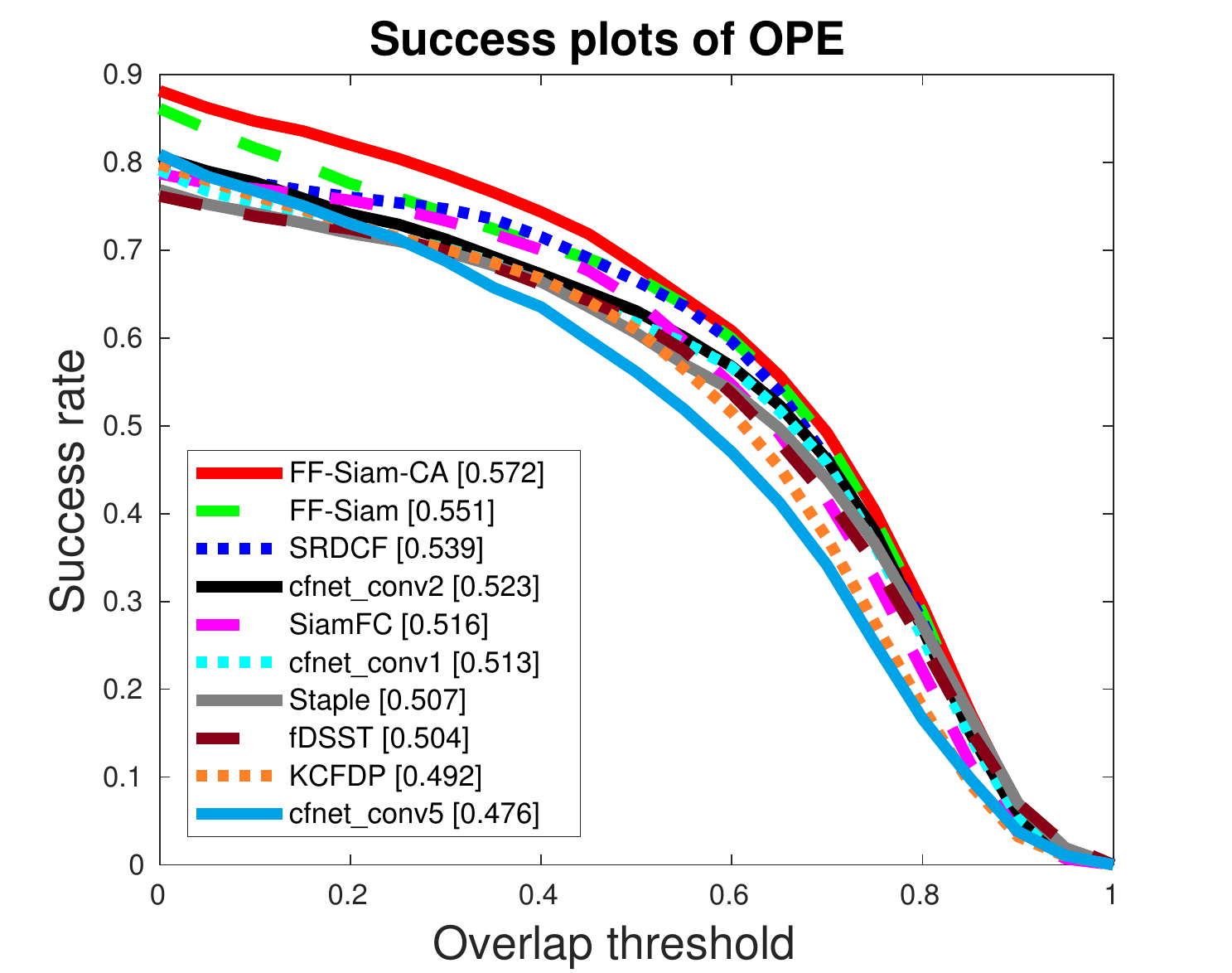}}
	\hspace{0.01\linewidth}
	\subfigure[OTB50]{\label{fig:UAV123pre}
		\includegraphics[width=0.4\linewidth,height=4.2cm]{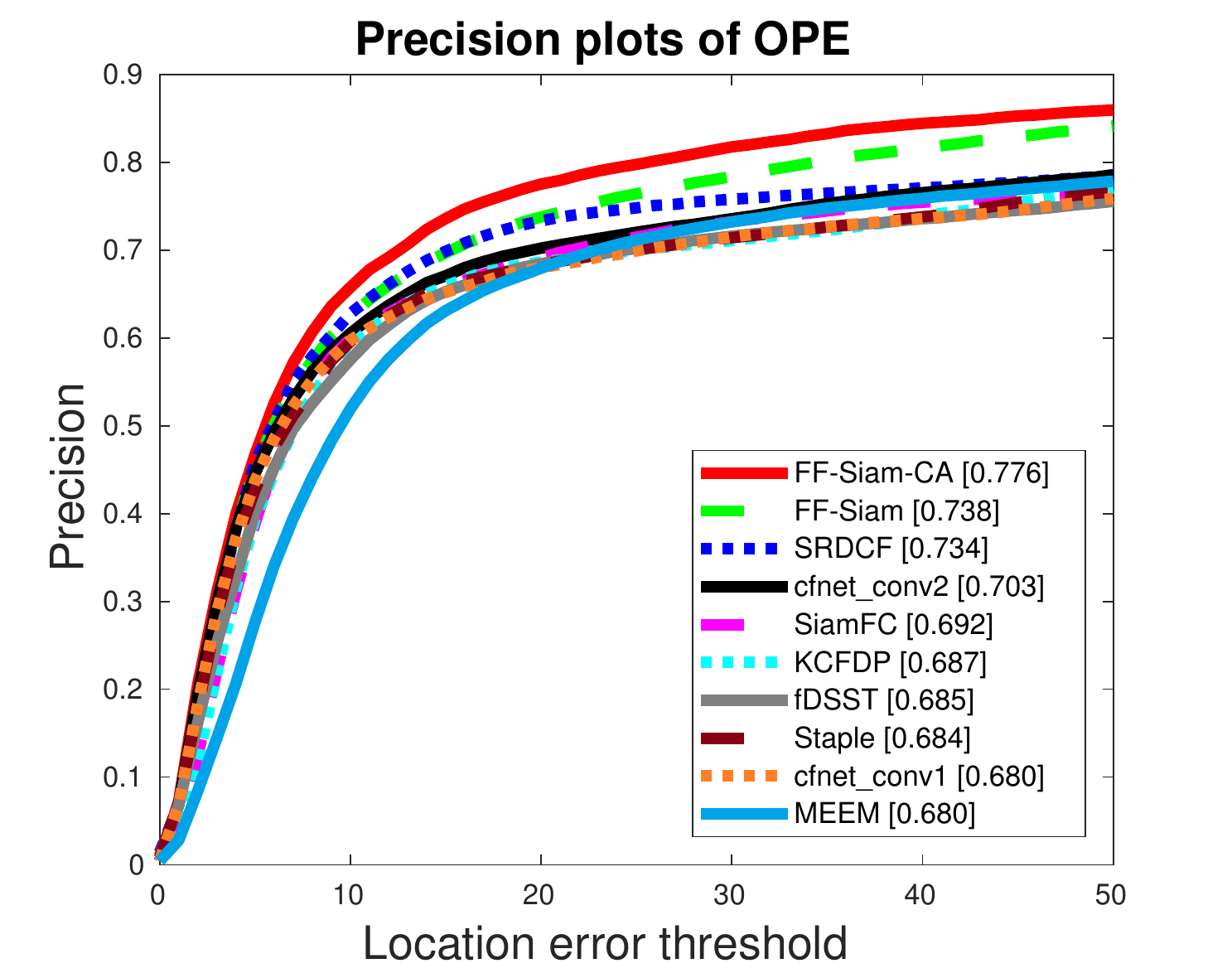}}
	\caption{Success plots on the UAV-123 (a) and OTB50 (c) datasets. Precision plots on the UAV123 (b) and OTB50 (d) datasets. The score of each tracker is shown in the legend. Our approach achieves state-of-the-art performance in all datasets. For clarity, only the results of top 10 trackers are shown in the legend.}
	\label{comparestate}
\end{figure}
\textbf{Analysis:} Figure 6 shows the results of precision and success rate respectively in UAV123 and OTB50. Figure 6(a) and 6(c) show the results of the success rate as described above, curves with different color show the results of different trackers. Figure 6(b) and 6(d) show the result of the precision rate which represents the percentage of frames with a center location error under the threshold. A frame whose distance is less than the threshold is considered to be accurate. The center location error threshold with 20 pixels is taken as the standard. Among the comparisons with the Siamese Network based approaches, our approach provides better performance than SiamFC and staple. Staple only fuses HOG features and color features, while SiamFC only use the CNN features. It proves that the semantic features and appearance features can complement each other, and the approach we proposed to fuse multiple features is effective. Besides, it shows that the FF-Siam$\_$CA which combined with channel attention get the 57.2\% score of AUC better than the FF-Siam with 55.1\%. This is because in different scenes, the performance of different characteristics are different, and the fusion weight of the feature channels should be changed. For example, when there are distractors with the same target category in search area, the semantic information is not the most important information. For this case, we should pay more attention to other features, so the weight of feature channels which contain semantic information should be decreased.

Figure 7 provides some visualization result for tracking. It is obvious that the proposed approach is effective for small targets which are difficult to  track. In the top, when occlusion occurs, most of the trackers have lost the target while the proposed tracker still work well. As shown in the middle , the proposed tracker is much more stable than others when the target becomes too small to be tracked. And as shown in the bottom, when the scale of the tracking target varies quickly, the proposed tracker adapts to fit the object while the others fails.

\begin{figure}
		\centering
	\includegraphics[width = \linewidth]{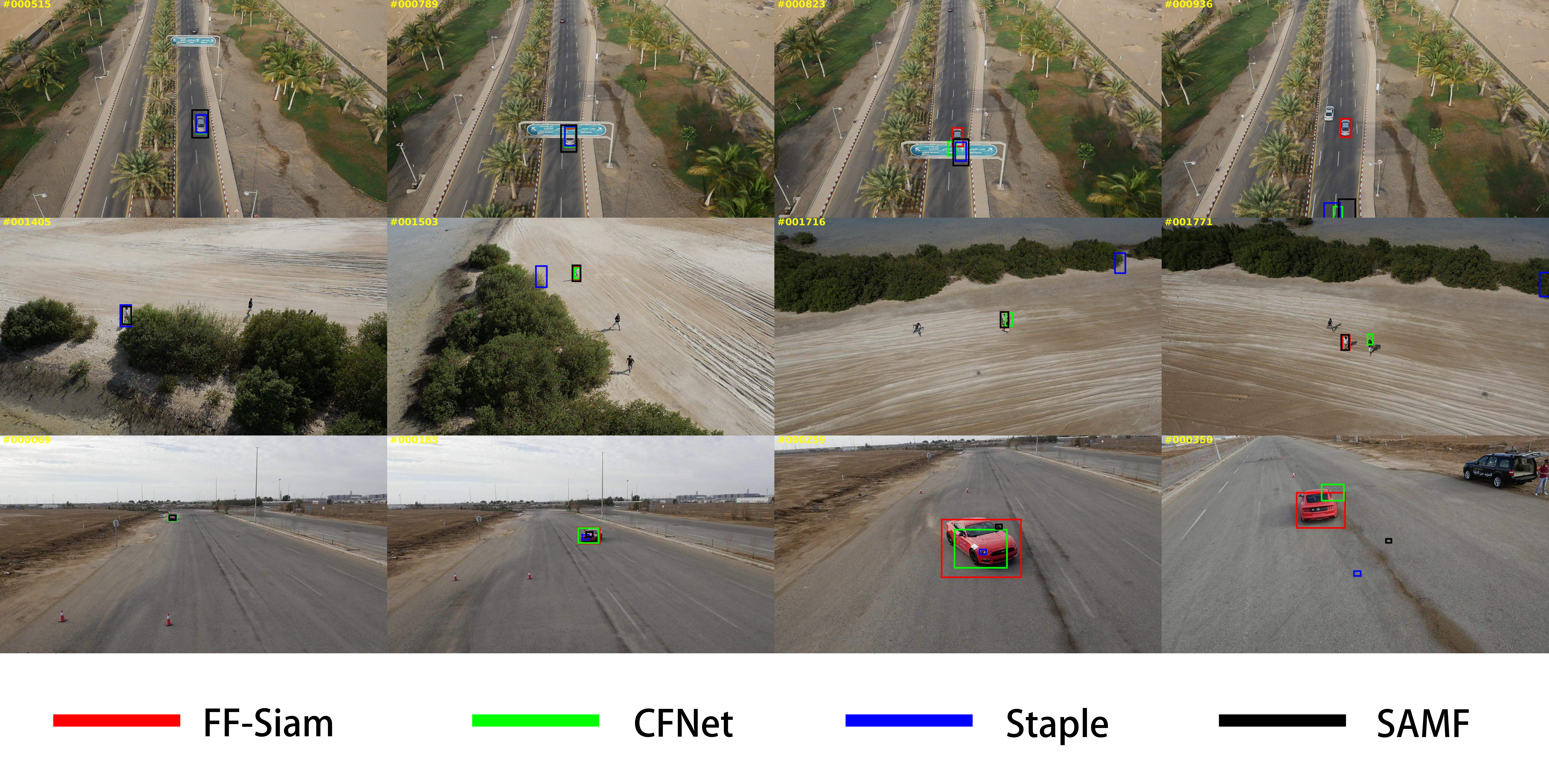}
	\caption{A qualitative comparison of the proposed approach with other three state-of-the-art approaches on three sequences: car9$\_$1 (as shown in the top), group2$\_$2 (as shown in the middle ) and car16$\_$1 (as shown at the bottom).}
	\label{fig:moreresult}
	
\end{figure}

\section{Conclusion}
In this paper, a novel end-to-end feature fusion approach is proposed based on Siamese Network for adaptive robust visual object tracking. The training of the network model makes up the defect of different features in the tracking effect. The proposed feature fusion network improves the generality of the tracker, achieving excellent performance in scenes with fast motion, motion blur, background clutters.\\

\section{Acknowledgment}
This work was supported in part by Natural Science Foundation of Zhejiang Province (LQ18F030013, LQ18F030014, LQ16F030007), inpart by the NSFC under Grant 61802348 and Innovation Foundation from Key Laboratory of Intelligent Perception and Systems for High-Dimensional Information of Ministry of Education (JYB201706).

\section*{References}
 \bibliography{bib_new}

\begin{thebibliography}{10}
\expandafter\ifx\csname url\endcsname\relax
  \def\url#1{\texttt{#1}}\fi
\expandafter\ifx\csname urlprefix\endcsname\relax\def\urlprefix{URL }\fi
\expandafter\ifx\csname href\endcsname\relax
  \def\href#1#2{#2} \def\path#1{#1}\fi

\bibitem{wu2015object}
Y.~Wu, J.~Lim, M.-H. Yang, Object tracking benchmark, TPAMI 37~(9) (2015)
  1834--1848.

\bibitem{danelljan2014accurate}
M.~Danelljan, G.~H{\"a}ger, F.~Khan, M.~Felsberg, Accurate scale estimation for
  robust visual tracking, in: BMVC, 2014.

\bibitem{danelljan2017discriminative}
M.~Danelljan, G.~H{\"a}ger, F.~S. Khan, M.~Felsberg, Discriminative scale space
  tracking, TPAMI 39~(8) (2017) 1561--1575.

\bibitem{zhang2014meem}
J.~Zhang, S.~Ma, S.~Sclaroff, Meem: robust tracking via multiple experts using
  entropy minimization, in: ECCV, 2014.

\bibitem{bertinetto2016staple}
L.~Bertinetto, J.~Valmadre, S.~Golodetz, O.~Miksik, P.~H. Torr, Staple:
  Complementary learners for real-time tracking, in: CVPR, 2016.

\bibitem{li2014scale}
Y.~Li, J.~Zhu, A scale adaptive kernel correlation filter tracker with feature
  integration., in: ECCV Workshops, 2014.

\bibitem{danelljan2015convolutional}
M.~Danelljan, G.~Hager, F.~Shahbaz~Khan, M.~Felsberg, Convolutional features
  for correlation filter based visual tracking, in: ICCV Workshops, 2015.

\bibitem{danelljan2016beyond}
M.~Danelljan, A.~Robinson, F.~S. Khan, M.~Felsberg, Beyond correlation filters:
  Learning continuous convolution operators for visual tracking, in: ECCV,
  2016.

\bibitem{ma2015hierarchical}
C.~Ma, J.-B. Huang, X.~Yang, M.-H. Yang, Hierarchical convolutional features
  for visual tracking, in: ICCV, 2015.

\bibitem{bertinetto2016fully}
L.~Bertinetto, J.~Valmadre, J.~F. Henriques, A.~Vedaldi, P.~H. Torr,
  Fully-convolutional siamese networks for object tracking, in: ECCV, 2016.

\bibitem{chen2017once}
K.~Chen, W.~Tao, Once for all: a two-flow convolutional neural network for
  visual tracking, TCSVT PP~(99) (2017) 1--1.

\bibitem{held2016learning}
D.~Held, S.~Thrun, S.~Savarese, Learning to track at 100 fps with deep
  regression networks, in: ECCV, 2016.

\bibitem{leal2016learning}
L.~Leal-Taix{\'e}, C.~Canton-Ferrer, K.~Schindler, Learning by tracking:
  Siamese cnn for robust target association, in: CVPR Workshops, 2016.

\bibitem{tao2016siamese}
R.~Tao, E.~Gavves, A.~W. Smeulders, Siamese instance search for tracking, in:
  CVPR, 2016.

\bibitem{henriques2015high}
J.~F. Henriques, R.~Caseiro, P.~Martins, J.~Batista, High-speed tracking with
  kernelized correlation filters, TPAMI 37~(3) (2015) 583--596.

\bibitem{valmadre2017end}
J.~Valmadre, L.~Bertinetto, J.~F. Henriques, A.~Vedaldi, P.~H. Torr, End-to-end
  representation learning for correlation filter based tracking, in: CVPR,
  2017.

\bibitem{bromley1993signature}
J.~{Bromley}, J.~W. {Bentz}, L.~{Bottou}, L.~{Bottou}, I.~{Guyon}, Y.~{Lecun},
  C.~{Moore}, E.~{Säckinger}, R.~{Shah}, Signature verification using a
  siamese' time delay neural network, International Journal of Pattern
  Recognition and Artificial Intelligence 7~(4) (1993) 669--688.

\bibitem{chopra2005learning}
S.~{Chopra}, R.~{Hadsell}, Y.~{LeCun}, Learning a similarity metric
  discriminatively, with application to face verification, in: CVPR, Vol.~1,
  2005, pp. 539--546.

\bibitem{taigman2014deepface}
Y.~{Taigman}, M.~{Yang}, M.~{Ranzato}, L.~{Wolf}, Deepface: Closing the gap to
  human-level performance in face verification, in: CVPR, 2014, pp. 1701--1708.

\bibitem{han2015matchnet}
X.~{Han}, T.~{Leung}, Y.~{Jia}, R.~{Sukthankar}, A.~C. {Berg}, Matchnet:
  Unifying feature and metric learning for patch-based matching, in: CVPR,
  2015, pp. 3279--3286.

\bibitem{zagoruyko2015learning}
S.~{Zagoruyko}, N.~{Komodakis}, Learning to compare image patches via
  convolutional neural networks, in: CVPR, 2015, pp. 4353--4361.

\bibitem{krizhevsky2012imagenet}
A.~{Krizhevsky}, I.~{Sutskever}, G.~E. {Hinton}, Imagenet classification with
  deep convolutional neural networks, in: NIPS, 2012, pp. 1097--1105.

\bibitem{wang2014ensemble}
N.~Wang, D.-Y. Yeung, Ensemble-based tracking: Aggregating crowdsourced
  structured time series data, in: ICML, 2014.

\bibitem{kwon2010visual}
J.~{Kwon}, K.~M. {Lee}, Visual tracking decomposition, in: CVPR, 2010, pp.
  1269--1276.

\bibitem{possegger2015defense}
H.~Possegger, T.~Mauthner, H.~Bischof, In defense of color-based model-free
  tracking, in: CVPR, 2015.

\bibitem{olshausen1994neurobiological}
B.~A. {Olshausen}, C.~H. {Anderson}, D.~V. {Essen}, A neurobiological model of
  visual attention and invariant pattern recognition based task.

\bibitem{hu2018squeeze}
J.~{Hu}, L.~{Shen}, G.~{Sun}, Squeeze-and-excitation networks, computer vision
  and pattern recognition.

\bibitem{jaderberg2015spatial}
M.~{Jaderberg}, K.~{Simonyan}, A.~{Zisserman}, K.~{Kavukcuoglu}, Spatial
  transformer networks, neural information processing systems (2015)
  2017--2025.

\bibitem{bahdanau2015neural}
D.~{Bahdanau}, K.~{Cho}, Y.~{Bengio}, Neural machine translation by jointly
  learning to align and translate, international conference on learning
  representations.

\bibitem{luong2015effective}
T.~{Luong}, H.~{Pham}, C.~D. {Manning}, Effective approaches to attention-based
  neural machine translation, in: Proceedings of the 2015 Conference on
  Empirical Methods in Natural Language Processing, 2015, pp. 1412--1421.

\bibitem{shan2017attention}
C.~{Shan}, J.~{Zhang}, Y.~{Wang}, L.~{Xie}, Attention-based end-to-end speech
  recognition in mandarin.

\bibitem{chorowski2014end}
J.~{Chorowski}, D.~{Bahdanau}, K.~{Cho}, Y.~{Bengio}, Y.~{Bengio}, Y.~{Bengio},
  End-to-end continuous speech recognition using attention-based recurrent nn:
  First results, arXiv preprint arXiv:1412.1602.

\bibitem{liang2015encoding}
P.~Liang, E.~Blasch, H.~Ling, Encoding color information for visual tracking:
  Algorithms and benchmark, TIP 24~(12) (2015) 5630--5644.

\bibitem{mueller2016benchmark}
M.~Mueller, N.~Smith, B.~Ghanem, A benchmark and simulator for uav tracking,
  in: ECCV, 2016.

\bibitem{wu2013online}
Y.~Wu, J.~Lim, M.-H. Yang, Online object tracking: A benchmark, in: CVPR, 2013.

\bibitem{danelljan2015learning}
M.~Danelljan, G.~Hager, F.~Shahbaz~Khan, M.~Felsberg, Learning spatially
  regularized correlation filters for visual tracking, in: CVPR, 2015.

\bibitem{danelljan2014adaptive}
M.~Danelljan, F.~Shahbaz~Khan, M.~Felsberg, J.~Van~de Weijer, Adaptive color
  attributes for real-time visual tracking, in: CVPR, 2014.

\bibitem{gao2014transfer}
J.~Gao, H.~Ling, W.~Hu, J.~Xing, Transfer learning based visual tracking with
  gaussian processes regression, in: ECCV, 2014.

\bibitem{huang122015enable}
D.~Huang, L.~Luo, M.~Wen, Z.~Chen, C.~Zhang, Enable scale and aspect ratio
  adaptability in visual tracking with detection proposals, in: BMVC, 2015.

\end{thebibliography}

\end{document}